\pdfoutput=1
\documentclass[runningheads]{llncs} %
\usepackage{graphicx,subcaption}
\usepackage{amsmath}
\usepackage{adjustbox}
\usepackage{caption}

\usepackage{enumitem}
\setlist{nosep} 
\setlist{itemsep=1pt, topsep=3pt}

\usepackage{array}
\newcolumntype{x}[1]{>{\centering\let\newline\\\arraybackslash\hspace{0pt}}p{#1}}

\begin{document}
\title{Untrimmed Action Anticipation}
%
%


\author{Ivan Rodin\inst{1,2} \and
Antonino Furnari\inst{1,3} \and
Dimitrios Mavroeidis\inst{2} \and\\
Giovanni Maria Farinella\inst{1,3}}

\authorrunning{I. Rodin et al.}
%
\institute{University of Catania,
Viale Andrea Doria 6, Catania, Italy, 95128\\
\email{\{ivan.rodin, antonino.furnari, giovanni.farinella\}@unict.it} \and
Philips Research, High Tech Campus 34, 5656 AE, Eindhoven, The Netherlands\\
\email{dimitrios.mavroeidis@philips.com} \and
Next Vision s.r.l – Spinoff of the University of Catania - www.nextvisionlab.it \email{info@nextvisionlab.it}}
\maketitle              
\begin{abstract}
Egocentric action anticipation consists in predicting a future action the camera wearer will perform from egocentric video. While the task has recently attracted the attention of the research community, current approaches assume that the input videos are ``trimmed'', meaning that a short video sequence is sampled a fixed time before the beginning of the action. We argue that, despite the recent advances in the field, trimmed action anticipation has a limited applicability in real-world scenarios where it is important to deal with ``untrimmed'' video inputs and it cannot be assumed that the exact moment in which the action will begin is known at test time. To overcome such limitations, we propose an untrimmed action anticipation task, which, similarly to temporal action detection, assumes that the input video is untrimmed at test time, while still requiring predictions to be made before the actions actually take place. We design an evaluation procedure for methods designed to address this novel task, and compare several baselines on the EPIC-KITCHENS-100 dataset. Experiments show that the performance of current models designed for trimmed action anticipation is very limited and more research on this task is required.

\end{abstract}
\section{Introduction}
\label{sec:introduction}


\begin{figure}
  \centering
  \begin{subfigure}[t]{.47\textwidth}
    \centering\includegraphics[width=\textwidth]{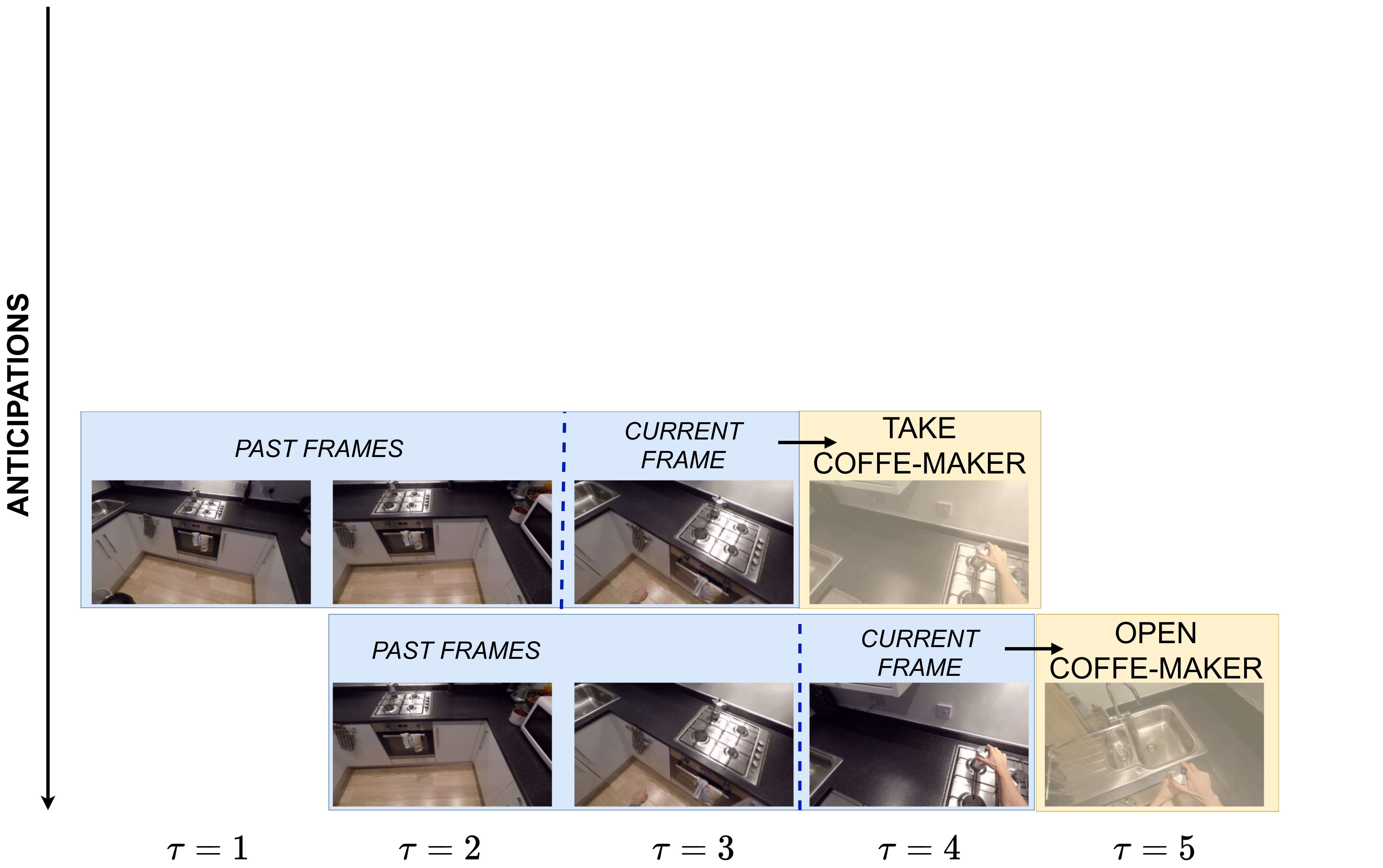}
    \caption{Trimmed}
  \end{subfigure}\qquad
  \begin{subfigure}[t]{.47\textwidth}
    \centering\includegraphics[width=\textwidth]{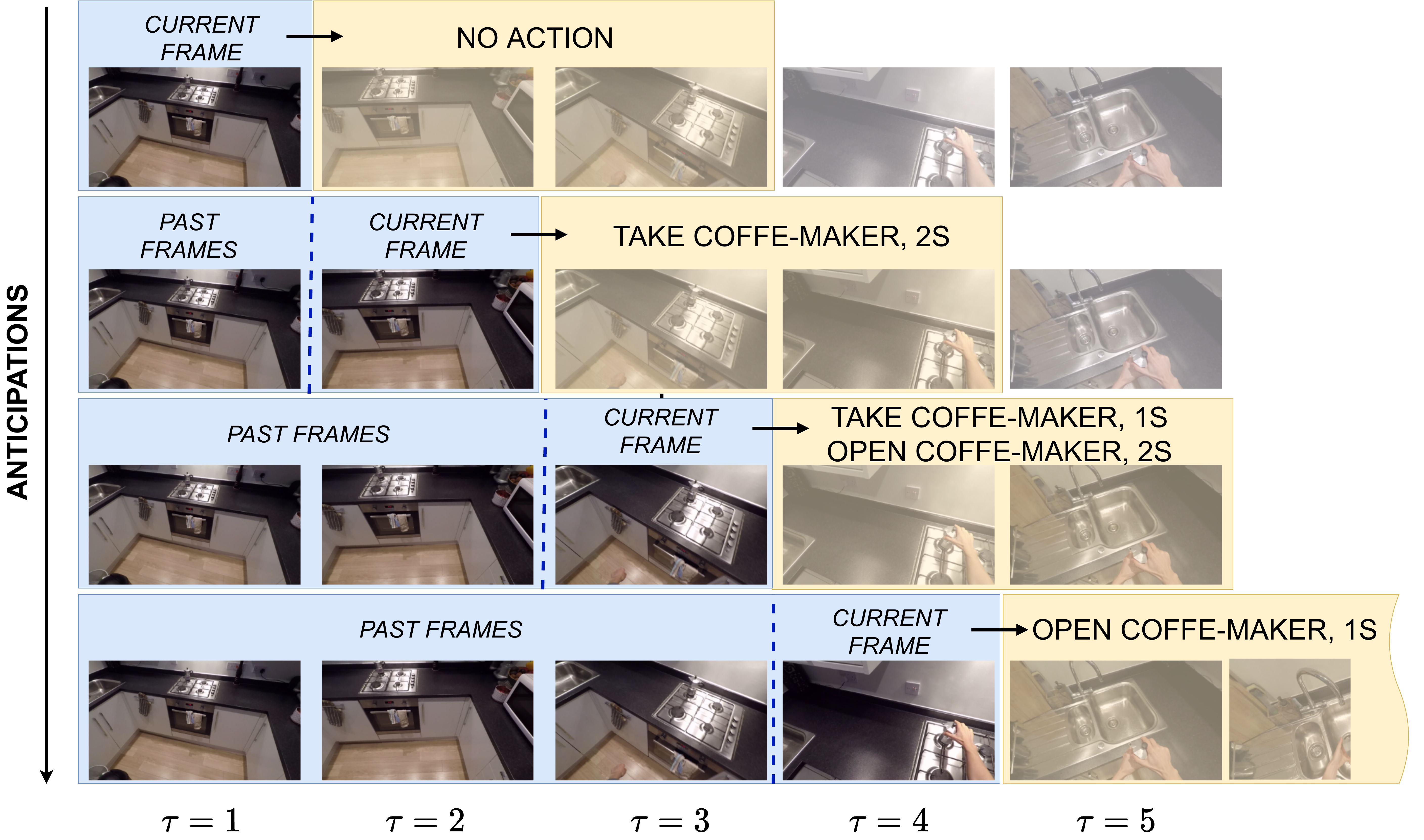}
    \caption{Untrimmed}
  \end{subfigure}
\caption{\small{Comparison of trimmed and untrimmed action anticipation. Trimmed action anticipation assumes that a video sequence is sampled prior to the beginning of the action with a fixed anticipation time (usually set to 1 second). Untrimmed action anticipation processes video frames consequently and predicts all the actions beginning within a reasonable temporal horizon along with time-to-action estimates, or determine that no activities will be performed.}}
\label{fig:tr_vs_untr}
\end{figure}

Egocentric or first-person vision concerns the understanding of videos captured by wearable devices such as action cameras or smart glasses \cite{betancourt2015evolution}. Egocentric videos provide visual observations from the unique point of view of the camera wearer, capturing hands, manipulated objects, head movements etc., which are useful to infer the behaviour of the users. The first-person point of view allows not only to detect and recognise the activities performed by a camera wearer, but also to model future intentions and activities \cite{rodin2021predicting}.
The egocentric action anticipation task, in particular, consists in predicting an action performed by the camera wearer before it occurs \cite{damen2018scaling}. Following the notation proposed in \cite{furnari2019would}, the task is to predict the class $y$ of an action starting at time $\tau_s$, by observing a \textit{trimmed} video segment preceding the action, i.e., a video segment starting at time $\tau_s-(\tau_o+\tau_a)$ and ending at time $\tau_s-\tau_a$. Here, $\tau_o$ is the length of the observed trimmed video and $\tau_a$ is the anticipation time, denoting how many seconds in advance  the anticipation should be performed. 
The above defined task, later referred to as ``trimmed action anticipation'' (Fig. \ref{fig:tr_vs_untr}(a)), has important limitations that restrict its applicability to real-life scenarios.

The first limitation arises from the observation that real action anticipation systems should be able to process video in an on-line fashion. More specifically, algorithms should be required to anticipate the occurrence of future actions at any time-step without knowing when the action will actually begin. We refer to this last task as ``untrimmed action anticipation''. For instance, \textit{untrimmed} processing capabilities would be beneficial in such applications as guiding blind people by modelling their intentions and trajectories \cite{ohn2018personalized}, or for human-robot interaction understanding \cite{koppula2015anticipating,ryoo2015robot}, in which case a continuous video stream is expected as input to the activity anticipation module. In this paper, we show that methods targeted to trimmed action anticipation (e.g., \cite{furnari2019would}) achieve limited performance in the untrimmed anticipation scenario.

The second issue limiting the applicability of trimmed action anticipation in real scenarios is the fixed anticipation time $\tau_a$, usually set to one second, which forces the action anticipation module to predict an action that will happen exactly $\tau_a$ seconds after the observed video segment. Indeed, setting a fixed anticipation time parameter $\tau_a$ does not follow the natural ability of humans to predict the future \cite{bubic2010prediction}. Rather than answering the question ``what will happen next and when?'', trimmed action anticipation answers the question ``what will happen after $\tau_a$ seconds?''. Taking in mind the aforementioned considerations, it would be reasonable for an anticipation system to attempt to predict the pair $\widehat{y_a} = (\widehat{c_a}, \widehat{\tau_a})$ consisting of the next action $\widehat{c_a} = (verb_a, noun_a)$ to answer the ``what'' question, as well as the time-to-action $\widehat{\tau_a}$ to answer the ``when'' question.

A third issue is related to the constraint of the current definition of action anticipation to anticipate just one future action. Indeed, we believe that it is natural to predict, when possible, not only the next action $\widehat{c_{next}}$, which is the closest in the future to the current timestamp, but a sequence of future actions $\widehat{c_i}$ along with their respective time-to-action estimations $\widehat{\tau_i}$. This modified prediction scheme increases the amount of information returned by the system and can be important for modelling short, densely distributed actions. For instance, if the model misses the next very short action, it could still be able to handle the predictions of the remaining future actions, which is not evaluated in the current trimmed anticipation task.

In an attempt to overcome all of the above-mentioned problems, we propose an ``untrimmed action anticipation'' task (Fig. \ref{fig:tr_vs_untr}(b)), which assumes an untrimmed video as input and aims to predict at each time-step all actions beginning within a given temporal horizon (e.g., all actions beginning within $5$ seconds). We propose an evaluation scheme which accounts for the different properties of the considered task and experimentally compare several baseline which are based on current trimmed action anticipation approaches. Results highlight that the proposed task is challenging for current methods designed to tackle trimmed action anticipation, suggesting that more research on this task is required.

\section{Related works}\label{sec:related}

Our work is related to previous investigations in the field. In \cite{damen2018scaling}, the task of trimmed egocentric action anticipation is presented. Despite previous studies on action anticipation \cite{furnari2019would}, \cite{liu2019forecasting}, \cite{ke2019time}, \cite{sener2020temporal} previous works have not explicitly considered the untrimmed version of the task. Nevertheless, some authors have investigated aspects closely related to this work.



The authors of \cite{manglik2019forecasting} considered the problem of forecasting the time-to-collision between a suitcase-robot and a nearby pedestrian in a streaming video. The task aims to detect only collision events, thus the authors treat and evaluate it as a regression task --- predicting time-to-action at every timestamp.
In \cite{neumann2019future}, the task of single future event prediction is considered: if and when a future event will occur. Three metrics are used for evaluation: 1) Event prediction accuracy, assessing whether the event will or will not occur within a prediction window; 2) Time-to-event error, which is the average absolute prediction error between the point estimate of the time-to-event distribution and the ground true time; 3) Model Surprise, indicating the quality and the confidence of the probabilistic output of the model.

Even if not concerned with future prediction, the task of online action detection~\cite{li2016online} is relevant to our work.
The task aims to predict an action class for the current frame of the video, how much time has passed since the beginning of the action and when the action will end. Three evaluation metrics are used to assess performance: 1) F1-score, for which correct detections are calculated based on the intersection over union (IoU) between prediction and ground truth action intervals; 2) Start Localization Score (SL-Score) based on the relative distance between predicted and ground truth start times; 3) End Localization Score (EL-score), similar to SL-score but referred to the end of the action.
Another relevant approach is StartNet, introduced in \cite{gao2019startnet}, which was developed to address the problem of Online Detection of Action Start (ODAS), where action classes and their starting times are detected in untrimmed streaming video input. Since the start time is predicted after the moment of action start is observed, also this work is not focused on future predictions. However, the untrimmed setting of the addressed problem, makes this paper relevant to our work. In particular, we will adopt the evaluation protocol based on point-level average precision (p-AP), used in \cite{gao2019startnet} and proposed in \cite{shou2018online}. %

\section{Problem formulation}\label{sec:problem}

\begin{figure*}[t!]
\centering
\includegraphics[width=0.8\textwidth]{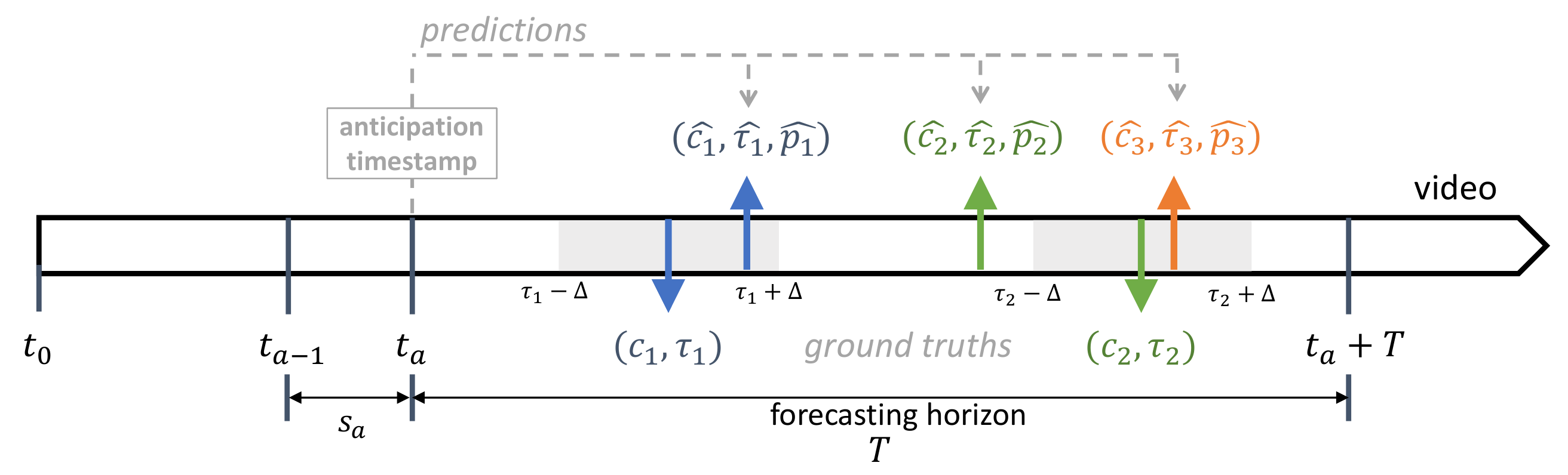}
\caption{\small{Untrimmed Action Anticipation. Anticipations are evaluated every $\alpha$ seconds. For the anticipation timestamp $t_a$, the figure illustrates two ground truth future actions $y_a = \{(c_i, \tau_i)\}_{i=1}^{2}$ (downwards arrows) and three predictions $\widehat{y_a} = \{(\widehat{c_i}, \widehat{\tau_i}, \widehat{s_i})\}_{i=1}^{3}$ (upwards arrows) within the anticipation horizon $T$. Grey segments denote the $\Delta$ temporal offset around the ground-truth actions (discussed in detail in Sec.~\ref{sec:evaluation}). Note that only $(\widehat{c_1}, \widehat{\tau_1})$ is a true positive prediction, since $(\widehat{c_2}, \widehat{\tau_2})$ does not lie within the temporal offset sampled around the ground truth action $(c_2, \tau_2)$, and $(\widehat{c_3}, \widehat{\tau_3})$ does not match the ground truth class label.}}
\label{fig:untrimmed_aa}
\end{figure*}


We define untrimmed action anticipation as the task of predicting the actions beginning within a given anticipation horizon, together with their starting times relative to the current timestamp (time-to-action). Since evaluating the task at every frame in a video can be redundant and computationally expensive, for each video, we choose a set of anticipation timestamps $t_1, t_2,..., t_n$ sampled every $\alpha$ seconds used for evaluation. For each anticipation timestamp $t_a$, we hence define a set of ground truth future actions $y_a$ to be anticipated, represented as follows: $y_a = \{(c_i, \tau_i)\}_{i=1}^{N_a}$, where $N_a$ is the number of future actions beginning within a fixed anticipation horizon $T$, $c_i$ is the class label of the $i$-th future action and $\tau_i\in [0, T]$ is the time-to-action.
The predictions at timestamp $t_a$ are presented in the form $\widehat{y_a} = \{(\widehat{c_i}, \widehat{\tau_i}, \widehat{s_i})\}_{i=1}^{N_C}$, where $N_C$ is the number of future actions predicted by the model, $\widehat{c_i}$ is the predicted class, $\widehat{\tau_i} \in \Re$ is the predicted time-to-action and $\widehat{s_i}\in[0,1]$ is the model's confidence about the predicted ``action, time-to-action'' $(\widehat{c_i}, \widehat{\tau_i})$ pair. 
The proposed untrimmed action anticipation task is illustrated in Figure~\ref{fig:untrimmed_aa}.



It is worth noting that we require models to predict a set of future actions, rather than just the next one as in trimmed action anticipation. This is a natural choice considering that, in our scenario, models predict both the future action class and its time-to-action. For instance, a model systematically skipping the next action but correctly predicting labels and time-to-action of the subsequent ones would be penalised in the classic evaluation scheme, whereas it is evaluated more fairly with the proposed protocol.

\section{Evaluation protocol}\label{sec:evaluation}


Since our task formulation involves two subtasks i.e., predicting the class labels of future actions (classification) and predicting the time-to-action values (regression), the evaluation metric should take into account both problems. More than that, the metric should account for the situations in which no future action is going to take place within the temporal horizon, and, in fact, these situations are a big part of datasets of natural activities as discussed in the experiments.
Considering all of the above, we propose to treat the untrimmed action anticipation task similarly to a (future) detection problem and define an evaluation protocol which has commonalities with the tasks of action localisation and object detection. 
Indeed, there are many similarities between untrimmed action anticipation and object detection or action localisation. Both tasks have to deal with many ``empty zones'' either in the timeline or in the image where no action or object is present. Similarly to untrimmed action anticipation, both of them combine classification (i.e., predicting the label of the action or object) and regression (i.e., predicting the beginning of the action or the coordinates of the bounding boxes). We hence base our evaluation measure on the popular mean Average Precision (mAP) defined in the Pascal VOC Object Detection Challenge \cite{everingham2010pascal} and adopted in many other works focused on object detection.

As in object detection, we first match each prediction to the most likely ground-truth future action annotation. Each prediction is then marked as either a true-positive or a false positive prediction. While in the Pascal VOC challenge the Intersection-Over-Union (IoU) was used to determine whether a detection is a true or a false positive by measuring the overlap between predicted and ground truth boxes, in untrimmed action anticipation, we propose to match a prediction to a ground truth action by measuring the temporal offset between the actual and predicted action start. Having the ground truth action $y_a$, its time-to-action $\tau_a$ and the predicted time-to-action $\hat{\tau_a}$, the temporal offset is calculated simply as $TO(\tau_a, \hat{\tau_a}) = |\tau_a - \hat{\tau_a}|$. The prediction $(\widehat{c_i}, \widehat{\tau_i}, \widehat{s_i})$ is considered as a true positive if the class $\widehat{c_i}$ matches the ground truth class $c_i$, and if the temporal offset is smaller than a specified threshold: $TO(\tau_i, \hat{\tau_i}) \leq \Delta$. An example of determining true or false positive predictions is illustrated in Fig.~\ref{fig:untrimmed_aa}.

Once predictions have been matched to the ground truth, we compute precision-recall (PR) curve for each action class from the list of the ranked predictions. Chosen a rank, the fraction of positive examples matched with predictions which are above the given rank is defined as the rank-based recall, whereas the proportion of the predictions above that rank which have been matched to a ground truth annotation is defined as the rank-based precision. Rank-based precision and recall values are computed for each possible rank in order to form a PR curve. The average precision per class is calculated as the mean interpolated precision with respect to a set of uniformly sampled recall levels with a step of 0.1: $[0, 0.1, ..., 1]$. Thus, $AP = \frac{1}{11}\sum_{r\in[0,0.1,...,1]} p_{interp}(r)$. Here, $p_{interp}(r)$ is the interpolated precision for the recall level $r$, computed by taking the maximum precision for which the corresponding recall exceeds $r$: $p_{interp}(r) = \max_{\widehat{r}:\widehat{r}\geq r}{p(\widehat{r})}$. The mean average precision ($mAP$) is obtained by averaging the $AP$ values obtained for each of the considered action classes.

For our benchmark, we compute mAP considering multiple temporal thresholds $\Delta \in [0.25, 0.5, 0.75, 1.0]$ seconds. We also explore a relaxed variation of the proposed metric for untrimmed action anticipation which computes mean average precision scores assuming an infinite temporal offset thresholding (mAP$_{\inf}$).

\section{Methods}\label{sec:experiments}

In this section, we report the experiments performed on the EPIC-KITCHENS-100 dataset (EK-100) \cite{damen2020rescaling}. We perform two types of experiments. First, we test the sensibility of the evaluation protocol sensibility with respect to the task. To do so, we conduct a set of controlled experiments in which we generate baseline results by randomly perturbing ground truth annotations with different amounts of randomness. Second, we evaluate different baselines for untrimmed action anticipation based on models designed for the trimmed task.

\subsection{Controlled experiments}
We propose a set of synthetic results obtained by randomly perturbing ground truth annotations using different perturbation parameters. These baselines are meant to provide a form of controlled experiments with predictors ranging from accurate to inaccurate based on the values chosen for the perturbation parameters. Our goal is to show that the proposed evaluation protocol does allow to estimate the model's ability to make good predictions. In these experiments, the random predictions $(\widehat{c_i}, \widehat{\tau_i}, \widehat{s_i})$ are generated from the ground-truth annotations $(c_i, \tau_i)$ as follows:

\begin{itemize}
    \item With a ``swapping'' probability $p_s$, the ground-truth class label $c_i$ is replaced with a label $\widehat{c_i}$ sampled from the class distribution in the training set:
    \begin{equation*}
    \widehat{c_i} = 
\begin{cases}
    c_s \sim EDF(c), &\text{if } x\leq p_s \\
    c_i,             &\text{otherwise}
\end{cases}
\end{equation*}
where $x\sim U(0,1)$, and $EDF(c)$ is the Empirical Distribution Function of action classes in the training set.
    \item The time-to-action $\widehat{\tau_i}$ is sampled from the normal distribution centred at the ground truth time-to-action with a pre-specified standard deviation $\sigma$: $\widehat{\tau_i} \sim N(\tau_i, \sigma)$
    \item The confidence score $\widehat{s_i}$ is set to $\widehat{s_i} = 1$
\end{itemize}

As a result, all action annotations $y_a = \{(c_i, \tau_i)\}_{i=1}^{N_a}$ are transformed to generated random predictions $\widehat{y} = (\widehat{c_{i}}, \widehat{\tau_{i}}, \widehat{s_{i}})_{i=1}^{N_a}$. For our experiments, we have explored different class swapping probabilities $x_s = [0.25, 0.5, 0.75]$ and different standard deviation values $\sigma = [0.33, 0.5, 1]$ for sampling time-to-action values.\looseness=-1

\subsection{Baselines based on trimmed action anticipation}
We propose different baselines for untrimmed action anticipation based on the Rolling Unrolling LSTM (RULSTM) model, which was originally designed for trimmed action anticipation \cite{furnari2019would}. The aim of these experiments is to explore the performance of current trimmed models when adapted to tackle the untrimmed action anticipation task. For simplicity, all these baselines rely on the RGB-inputs only.


\textbf{RU:} For the first baseline, we evaluate the original RULSTM model in an untrimmed manner, evaluating predictions every $0.25$ seconds. The original RULSTM model was trained to predict a future action one second before it occurs. Thus, we assume that the time-to-action for all the predictions is fixed to $\widehat{\tau_i} = 1 sec$. For action label predictions we select all the classes with confidence larger than the threshold $\widehat{s_i}>0.1$, providing the model with an opportunity to predict multiple future actions. If the highest action prediction confidence is smaller than or equal to the threshold $\widehat{s_i}\leq 0.1$ we assume that the model has predicted no action. We consider this baseline to assess the suitability of current models optimised for trimmed action detection in the considered untrimmed scenario.

\textbf{RU-no-act:} For this baseline, we investigate whether re-training the RULSTM model on all timestamps (rather than on timestamps sampled exactly 1s before each action) and adding the opportunity to predict a ``no action'' as an additional label can provide better performance.

\textbf{RU-sigmoid:} In this baseline, we replace the final SoftMax layer in the RULSTM architecture with a sigmoid layer, and train the model using binary cross-entropy, thus adjusting the original model for a multi-class multi-label classification task.

\textbf{RU-reg:} This baseline is aimed to predict the time-to-action $\widehat{\tau_i}$ by exploiting an additional fully-connected layer attached to the RULSTM model and trained to solve the regression task with mean squared error as a loss function.

\textbf{RU-5-clf:} In order to infer the time-to-action predictions, this baseline makes use of five separate RULSTM models trained to predict future actions at various time scales of 1, 2, 3, 4 and 5 seconds.

\section{Experimental settings and results}
\subsection{Dataset}\label{sec:dataset}

We perform our experiments on the EPIC-KITCHENS-100 dataset~\cite{damen2020rescaling}. To make the dataset suitable for untrimmed action anticipation, we sampled timestamps every $\alpha = 0.25$ seconds, leading to $1.057.238$ anticipation timestamps in the training set and $185.532$ anticipation timestamps in the validation set.

The statistics of the retrieved anticipation timestamps indicate two important features of the task. First, $38\%$ of anticipation timestamps annotations are not associated to future action labels within an anticipation horizon of five seconds, which highlights the necessity of modelling the \textit{``no action''} predictions in real scenarios. Second, $36\%$ of the anticipation timestamps contain at least two future action labels within the anticipation horizon, indicating the importance of being able to predict multiple future actions.

\label{sec:results}
\subsection{Results of the controlled experiments}

\begin{figure}[t]
  \centering
  \begin{subfigure}[t]{.29\textwidth}
    \centering\includegraphics[width=\textwidth]{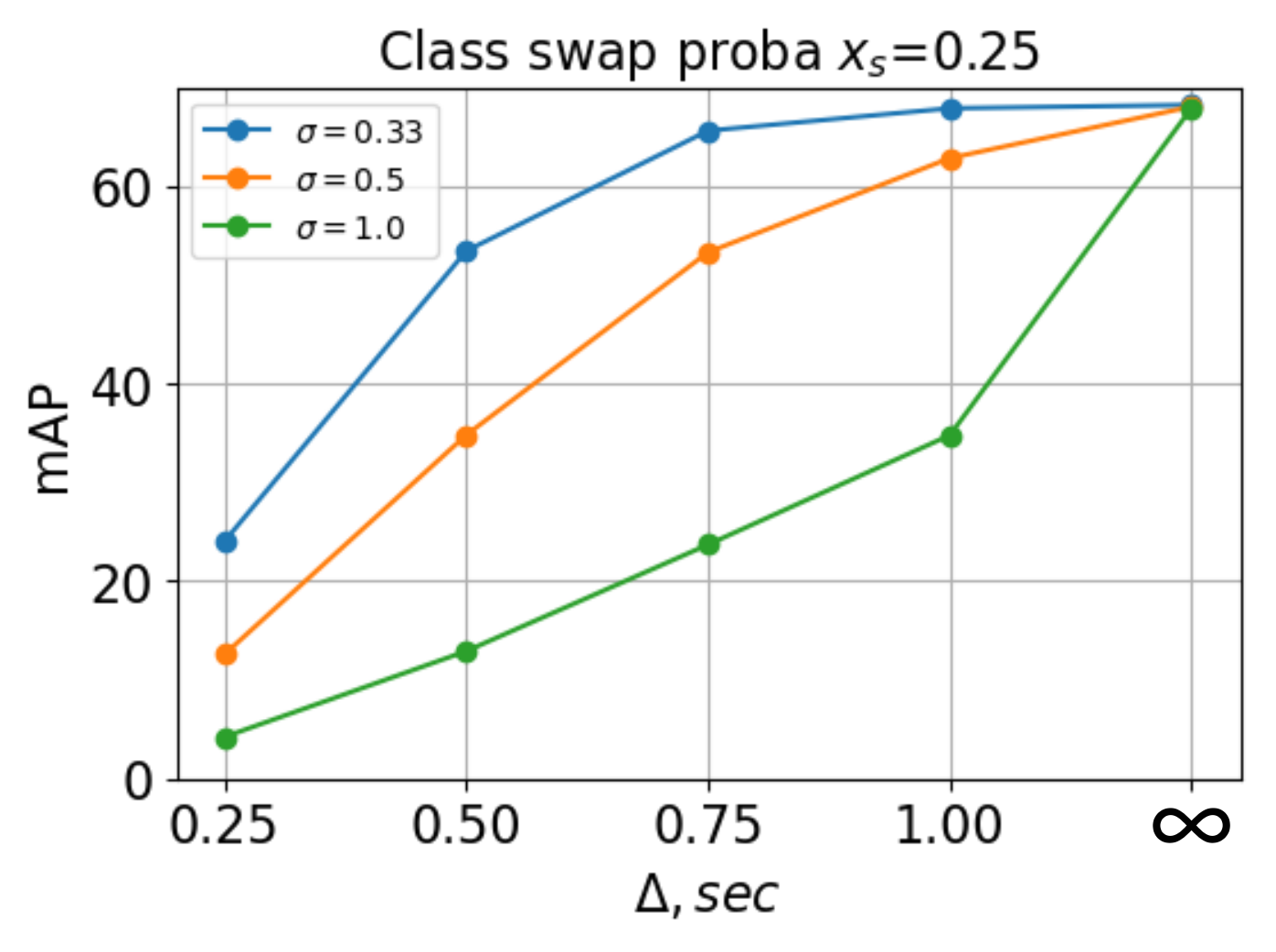}
  \end{subfigure}\qquad
  \begin{subfigure}[t]{.29\textwidth}
    \centering\includegraphics[width=\textwidth]{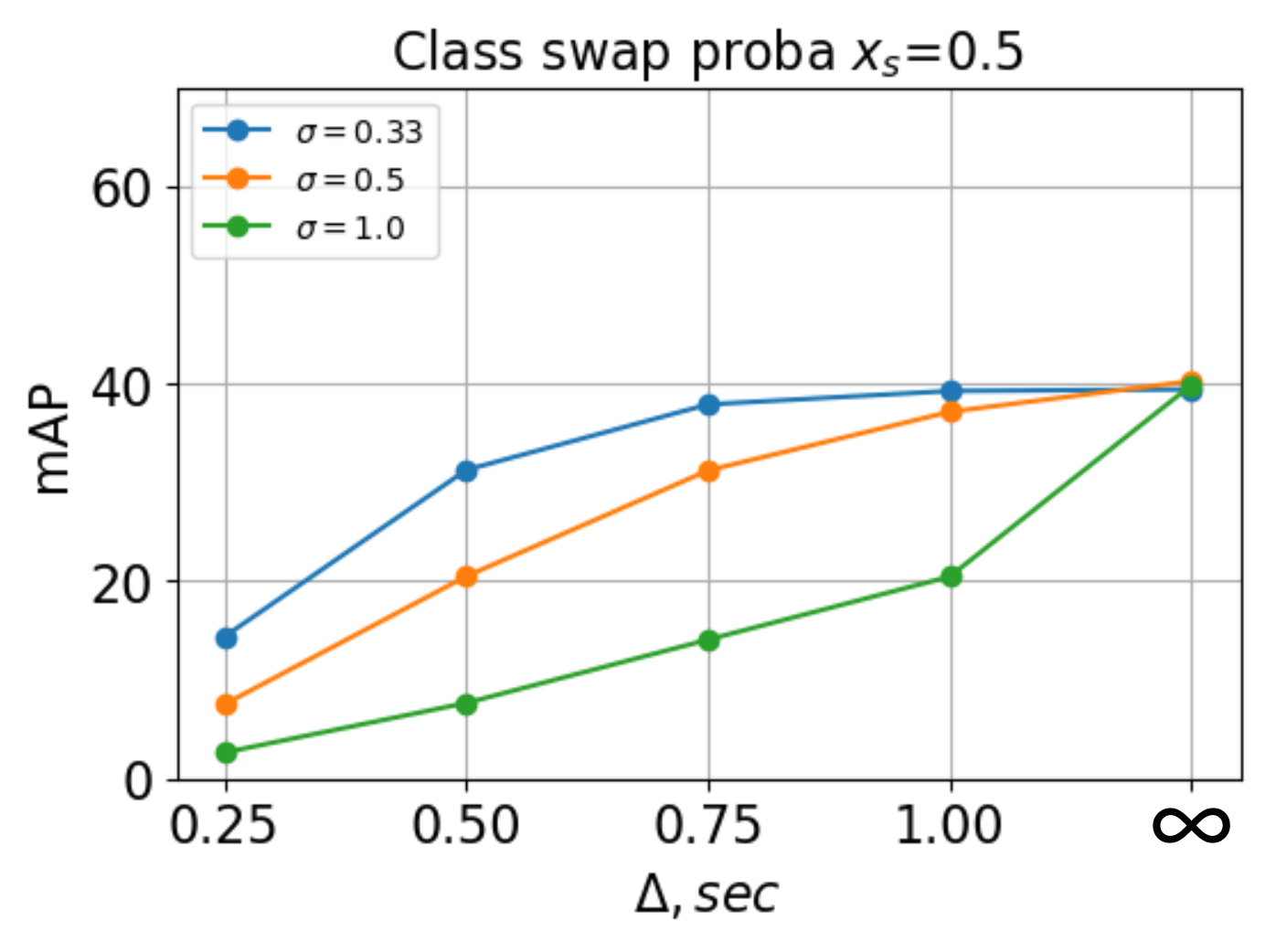}
  \end{subfigure}\qquad
  \begin{subfigure}[t]{.29\textwidth}
    \centering\includegraphics[width=\textwidth]{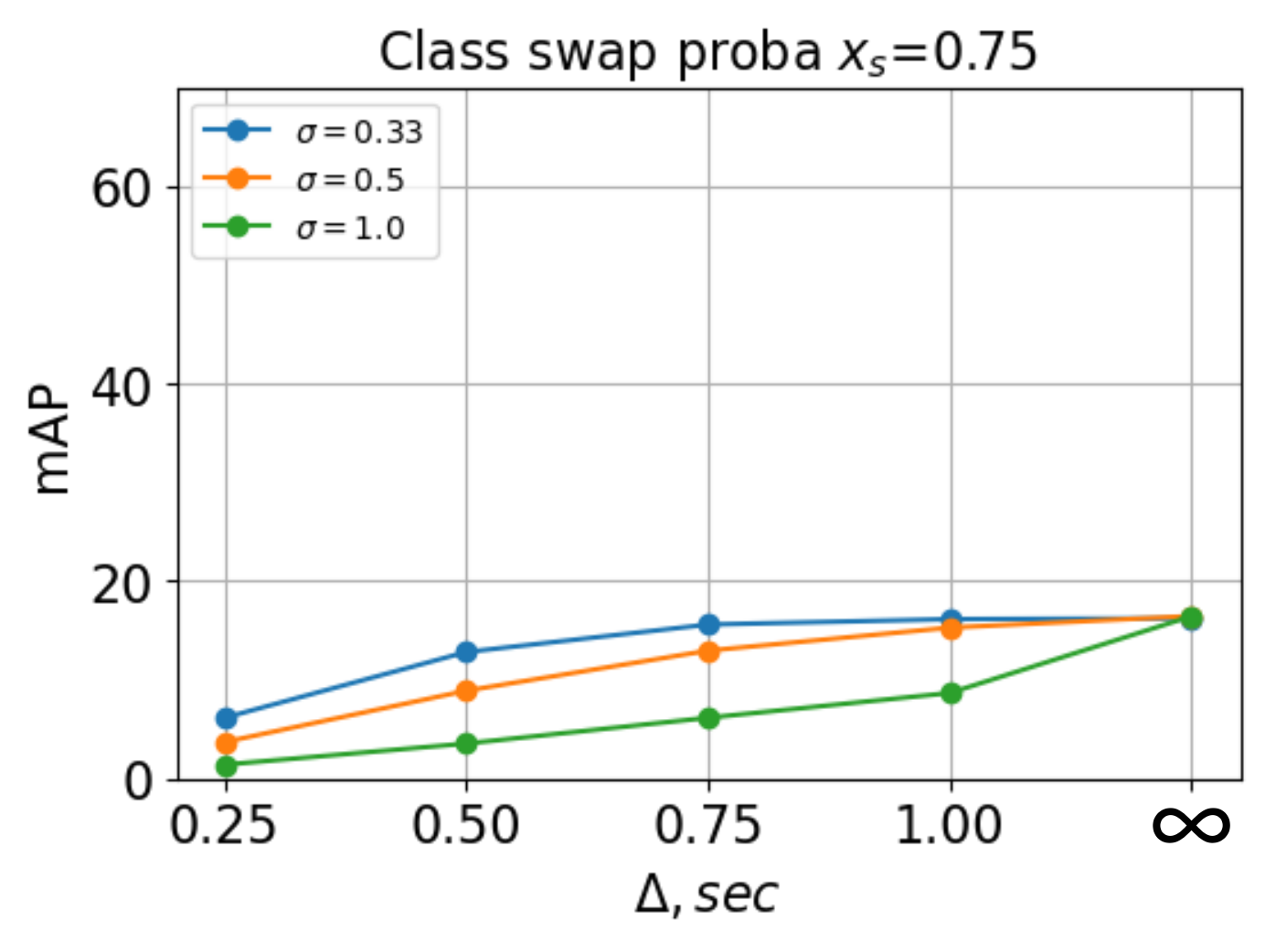}
  \end{subfigure}
  \caption{\small{Mean Average Precision Scores for the Random Baselines.}}
  \label{fig:genresults}
\end{figure}
Figure~\ref{fig:genresults} reports the results of the controlled experiments for different choices of the perturbation parameters. The results show that the proposed evaluation protocol is sensitive to the perturbations of the parameters. In particular, the higher the class swapping probability is, the lower the metric scores are (compare the different plots in Figure~\ref{fig:genresults}). Similar considerations apply to the perturbations of the time-to-action parameter: the larger the deviation from the ground truth time-to-action values of an action, the smaller the mAP score (compare the different series of the plots in Figure~\ref{fig:genresults}). Naturally, the mAP$_{\inf}$ score (shown for $\Delta=\infty$) reports the same results for different values of $\sigma$, since the relaxed version of the metric does not penalises the predictions in the presence of wrong time-to-action estimates.
Overall, the results suggest that the proposed evaluation protocol is suitable to evaluate models on this task. Indeed, a near-perfect method (blue line in the leftmost plot of Figure~\ref{fig:genresults}) would achieve reasonable performance ($mAP~30\%$) even for small values of the $Delta$ parameter. Nevertheless, the task is challenging, as shown by the results in the rightmost plot of Figure~\ref{fig:genresults}.

\subsection{Results for baselines based on trimmed action anticipation models}

\begin{table}[t]

\begin{minipage}{.49\linewidth}

\label{tab:rultsm_results}
\adjustbox{max width=\textwidth}{%

\begin{tabular}{|c|c|c|c|c|c|c|}
 \hline
 \multicolumn{6}{|c|}{Action} \\ \hline
 Model & mAP$_{0.25}$ & mAP$_{0.5}$ & mAP$_{0.75}$ & mAP$_{1.0}$ & mAP$_{\inf}$ \\
 \hline
 RU & \textbf{0.12} & \textbf{0.31} & \textbf{0.68} & \underline{0.84} & \underline{2.05}\\
 \hline
 RU-no-act & 0.07 & \underline{0.24} & 0.46 & 0.79 & 1.25\\
 \hline
 RU-sigmoid &0.01 & 0.02 & 0.03 & 0.06 & 0.11\\
  \hline
 RU-reg & 0.01 & 0.01 & 0.02 & 0.04 & 0.07\\
 \hline
 RU-5-clf & \underline{0.11} & \textbf{0.31} & \underline{0.66} & \textbf{0.87} & \textbf{2.55}\\

 \hline
 \hline
  \multicolumn{6}{|c|}{Verb} \\ \hline
 Model & mAP$_{0.25}$ & mAP$_{0.5}$ & mAP$_{0.75}$ & mAP$_{1.0}$ & mAP$_{\inf}$ \\
 \hline
 RU & \underline{0.25} & \underline{0.67} & \underline{1.28} & \underline{2.01} & \underline{4.82}\\
 \hline
 RU-no-act & 0.13 & 0.42 & 0.84 & 1.43 & 2.39\\
 \hline
 RU-sigmoid &0.02 & 0.03 & 0.05 & 0.10 & 0.19\\
 \hline
 RU-reg & 0.01 & 0.02 & 0.03 & 0.05 & 0.11\\
 \hline
 RU-5-clf & \textbf{0.27} & \textbf{0.69} & \textbf{1.34} & \textbf{2.15} & \textbf{5.30}\\
 \hline
 \hline
 \multicolumn{6}{|c|}{Noun} \\ \hline
 Model & mAP$_{0.25}$ & mAP$_{0.5}$ & mAP$_{0.75}$ & mAP$_{1.0}$ & mAP$_{\inf}$ \\
 \hline
 RU & \underline{0.23} & \underline{0.60} & \underline{1.14} & \underline{1.79} & \underline{4.25}\\
 \hline
 RU-no-act & 0.10 & 0.39 & 0.77 & 1.32 & 2.05\\
 \hline
 RU-sigmoid &0.02 & 0.03 & 0.04 & 0.09 & 0.18\\
 \hline
 RU-reg &0.01 & 0.01 & 0.02 & 0.04 & 0.10\\
 \hline
 RU-5-clf &\textbf{0.23} & \textbf{0.61} & \textbf{1.18} & \textbf{1.95} & \textbf{5.02}\\
 \hline
 
\end{tabular}
}
\captionof{table}{\small{Untrimmed anticipation results.}}

\end{minipage}
\hfill
\begin{minipage}{.49\linewidth}

\adjustbox{max width=\textwidth}{%
\begin{tabular}{|c|x{2.1cm}|x{2.1cm}|x{2.1cm}|}
 \hline
 \multicolumn{4}{|c|}{mAP$_{1.0}$, action} \\ \hline
 Model & untrimmed ts & untrimmed ts, excl. no-act & trimmed ts\\ \hline
 RU & \underline{0.84} & \underline{1.07} & \textbf{4.72} \\
 \hline
 RU-no-act & 0.79 & 1.02 & 2.98\\
 \hline
 RU-sigmoid & 0.06 & 0.09 & 0.29 \\
 \hline
 RU-reg & 0.04 & 0.07 & 0.17\\
 \hline
  RU-5-clf & \textbf{0.87} & \textbf{1.12} & \underline{3.17} \\
 \hline
\end{tabular}
}
\captionof{table}{\small{Results with respect to different sets of anticipation timestamps. ``untrimmed ts'': whole set of timestamps for the untrimmed task; ``untrimmed ts, excl. no-act'': all timestamps, excluding ``no future action''; ``trimmed ts'': timestamps and action instances from the trimmed task.}}

\end{minipage}

\end{table}

The results of the baselines based on trimmed action anticipation models are presented in Table~1. Best results are highlighted in bold, whereas second-best scores are underlined. 
The results highlight that the considered untrimmed action anticipation task is challenging. Indeed, the standard RU model obtains modest results for action, verb and noun even for the relaxed $mAP_{inf}$ metric ($2.05$, $4.82$ and $4.25$).
Results are worse if more realistic measures are considered (e.g., see the $mAP_{0.5}$ column.
Among the different baselines considered for comparisons, the combination of five RU models (RU-5-clf) consistently achieves best or second-best results.
Simply adding a ``no action'' class (RU-no-act), replacing a softmax with a sigmoid to enable multiple detections (RU-sigmoid) or adding a regression layer (RU-reg) generally lead to decreased performance, which further highlight the challenging nature of the task and the limits of current models optimised on trimmed action detection. %
We observe that the trimmed models tested in the untrimmed task show a very large number of false positive predictions for the ``no action'' anticipation timestamps. In particular, exploratory analysis of the results shows that the best performing model (RU-5-clf) generates false positive predictions for $98\%$ of the timestamps with no ground-truth action class, indicating the need of better ways to model the ``no action'' scenario appropriately.
It should also be noted that more attention should be paid to the time-to-action prediction. For all of the considered baselines the mAP$_{inf}$ score is significantly higher than the mAP$_{1.0}$ score, indicating that in some cases the models could predict future actions, but they could not estimate the correct time-to-action even when a big temporal offset of one second is used for evaluation.

Table~2 compares the performance of the baselines for different sets of anticipation timestamps when the $mAP_{1.0}, action$ measure is considered. The different sets include the untrimmed timestamps (same as the ones of Table~1), all timestamps excluding ``no action'' cases, and timestamps sampled exactly 1 second before the beginning of labeled actions (trimmed case).
The results indicate the tendency that the models adapted from the trimmed task to obtain worst mAP scores when processing anticipation timestamps with multiple future actions or without defined future actions. Indeed, the task is much easier when trimmed timestamps are considered (last column).


\begin{figure*}[t]
\centering
\small{(a)} \includegraphics[width=\textwidth]{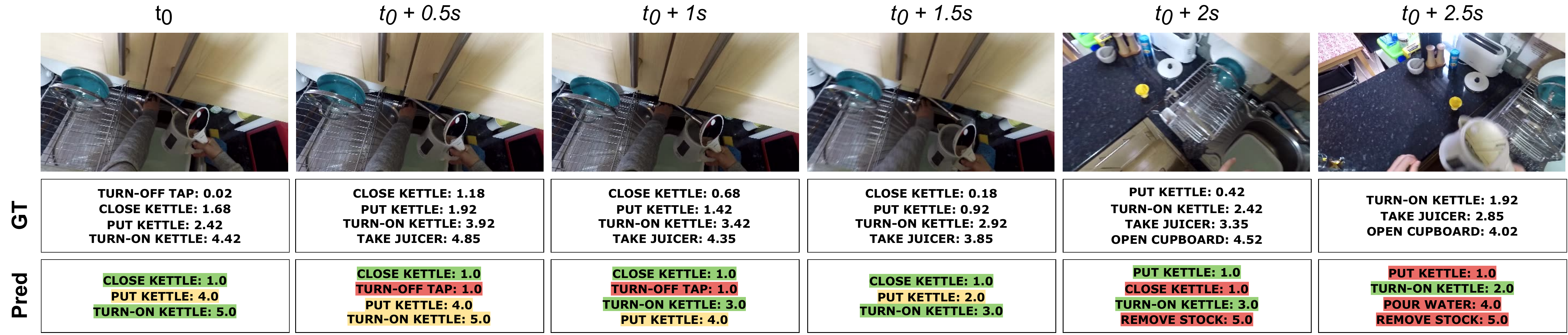}
\small{(b)} \includegraphics[width=\textwidth]{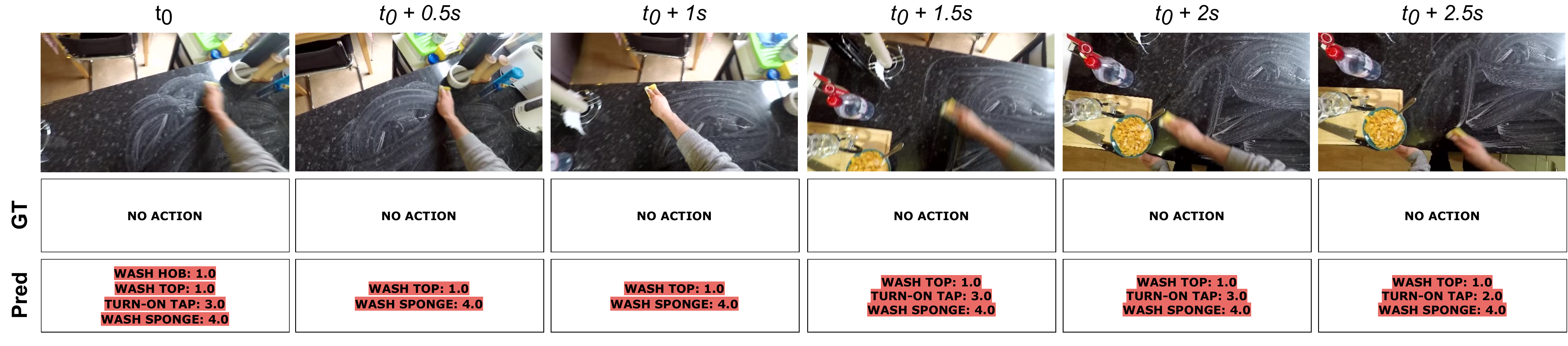}
\caption{\small{Example sequences from untrimmed videos along with GT annotations of future actions and the predictions of the RU-5-clf baseline. True positive predictions are highlighted in green(considering $\Delta=1s$). Predictions with correct class, but inaccurate time-to-action estimates are reported in yellow. Wrong predictions are reported in red.}}
\label{fig:qualitative}
\end{figure*}

The qualitative results of the experiments are presented in Figure~\ref{fig:qualitative}. The best performing model (RU-5-clf) is able to tackle the multiple actions prediction, as well as to predict the respected time-to-action estimates (Figure~\ref{fig:qualitative}a, frames 1, 3, 4, 5). Several predictions match with ground truth classes of future actions but have big error in time-to-action estimate (yellow), while sometimes the model is predicting the ongoing or past actions (\textit{turn-off tap} on frames 2 and 3). It is also hard for model to predict the actions which are not aligned with common intuition of the human activities, i.e., predicting the \textit{take juicer} action after filling the kettle with water, especially at the beginning when the juicer is out of the view (frames 2, 3, 4).

Much poorer performance is shown on the long sequences of the monotonic actions (Fig.~\ref{fig:qualitative}b). The model tends to predict the beginning of the new action(s) within the anticipation horizon, however, the ongoing action does not terminates for a long time, and all the inferred predictions become false positives.

\section{Conclusion}\label{sec:conclusion}

We have proposed a new task of untrimmed action anticipation. Our findings have indicated that modes optimised for trimmed anticipation have important limitations in this scenario, leading to many false-positive predictions and poor performance in the task of identifying the time-to-action. 
Retraining the same trimmed models on the video segments sampled with a sliding window, with the opportunities to predict the ``no action'' or multiple future actions also does not solve the problem if only the final layers of the model are changed. We believe more research is needed to build new solutions for the proposed task.
We would also like to emphasise the attention on terminology: we refer to the proposed task as \textit{untrimmed} action anticipation rather than \textit{streaming} action anticipation \cite{furnari2021towards}. For our problem formulation, the inference time does not play a primary role, hence we do not consider real-time evaluation.
Although we do not aim at improving the speed of the algorithms and do not consider real-time evaluation, we believe that streaming scenario is an important direction for future research in untrimmed action anticipation.

\section*{Acknowledgements}

This research has been supported by Marie Skłodowska-Curie Innovative Training Networks - European Industrial Doctorates - PhilHumans Project, European Union - Grant agreement 812882 (\url{http://www.philhumans.eu}), project MEGABIT - PIAno di inCEntivi per la RIcerca di Ateneo 2020/2022 (PIACERI) – linea di intervento 2, DMI - University of Catania, and by the MISE - PON I\&C 2014-2020 - Progetto ENIGMA - Prog n. F/190050/02/X44 – CUP: B61B19000520008.

%
%
%
\bibliographystyle{splncs04}
\bibliography{mybibliography}

%




\end{document}